# Generative Sensing: Transforming Unreliable Sensor Data for Reliable Recognition

Lina J. Karam, Tejas Borkar, Yu Cao, Junseok Chae
*School of ECEE, Arizona State University, Tempe, Arizona, USA*
{karam, tsborkar, Yu.Cao, Junseok.Chae}@asu.edu

## Abstract

*This paper introduces a deep learning enabled generative sensing framework which integrates low-end sensors with computational intelligence to attain a high recognition accuracy on par with that attained with high-end sensors. The proposed generative sensing framework aims at transforming low-end, low-quality sensor data into higher quality sensor data in terms of achieved classification accuracy. The low-end data can be transformed into higher quality data of the same modality or into data of another modality. Different from existing methods for image generation, the proposed framework is based on discriminative models and targets to maximize the recognition accuracy rather than a similarity measure. This is achieved through the introduction of selective feature regeneration in a deep neural network (DNN). The proposed generative sensing will essentially transform low-quality sensor data into high-quality information for robust perception. Results are presented to illustrate the performance of the proposed framework.*

## 1. Introduction

Recent advances in machine learning coupled with the accessibility of compact and low-cost multimedia sensors, such as visible image and infrared (IR) cameras, have fundamentally altered the way humans live, work and interact with each other. However, the performance of available sensors is still significantly constrained by the quality/cost tradeoff of the sensors themselves and by the sensitivity of deep learning algorithms to variations in sensor data quality and sensor modality. For example, image quality may be impacted by environmental factors, i.e., lighting conditions, and specifications of sensors, i.e., number of pixels, dynamic range, noise, etc. Under low-lighting conditions, the visible wavelength image suffers while the IR image may not largely suffer from the dark

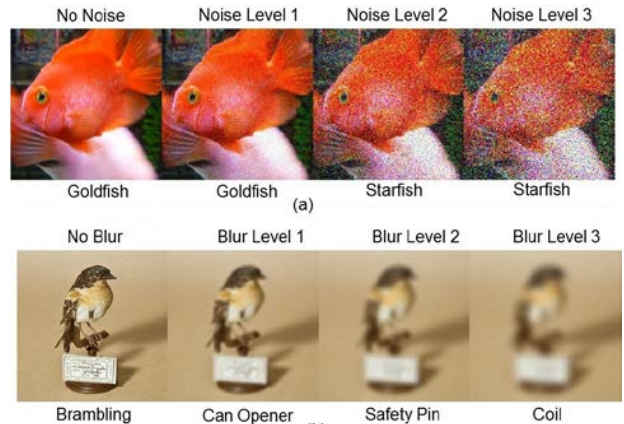

**Figure 1**. Effect of image quality on DNN predictions when using an AlexNet model [4] trained on high-resolution images. Degradation severity increases from left to right for white additive noise (a) and Gaussian blur (b). It can be seen that, while the quality degradation does not hinder the human ability to classify the images, the object class label predicted by the DNN changes significantly even at low degradation levels.

environment; the quality of both (visible and IR) may suffer due to low-end sensor performance.

Low-power low-profile (thin) sensors are often limited to produce low-resolution (blurry) and low SNR (noisy) images. Dodge and Karam [1] showed that such degradations, even at relatively small levels, can cause a significant drop in the classification performance of state-of-the-art DNNs which are typically trained using high-resolution images. Before the work of Dodge and Karam [1], the image quality was an aspect very often overlooked while designing deep learning based image classification systems. Dodge and Karam also showed very recently [2][3] that, although DNNs perform better than or on par with humans on pristine high-resolution images, the DNN classification performance is still significantly much lower than human performance on images with degradations in quality. Fig. 1 illustrates that, while humans are capable of recognizing with a reasonable accuracy objects in low-resolution blurred images or in low

SNR noisy images, DNNs that are trained on pristine images predict incorrect class labels even at a relatively low-level of perceivable degradation.

A visible wavelength image sensor, denoted as image sensor, is a core part of digital cameras and smart phones [5][6]. The cost of image sensors has been aggressively scaled largely due to the high-volume consumer market products. In addition to the cost, the technical specification of image sensors, including the number of pixels, color contrast, dynamic range, and power consumption, meet almost all demands of consumer market products. IR sensors, on the other hand, have been mostly used for specific needs such as surveillance- and tracking-based applications in the military domain, yet some started penetrating the consumer market recently [7][8]. IR sensors typically cost significantly higher than image sensors to produce IR images at an equivalent resolution primarily due to the relatively low volume market. Generally those sensors show a trend of "the higher the cost is, the better the delivered performance."

This work develops a deep learning based framework, which we refer to as *generative sensing*, that enables attaining the classification accuracy of a higher quality, high-end sensor while only using a low-quality, low-end sensor. Furthermore, the high-end sensor and low-end sensor can be of different types, in which case the proposed framework can be seen as transforming one type of sensor (e.g., near infrared (NIR) or IR sensor) into another type of sensor (e.g., visible wavelength image sensor) in terms of achieved classification performance. This is performed through feature regeneration for improved classification. This is important for enabling low-power and low-cost sensing platforms without compromising the recognition performance.

This paper is organized as follows. Section 2 discusses related work. Section 3 describes the proposed generative sensing framework. Illustrative performance results are presented in Section 4 and a conclusion is given in Section 5.

## 2. Related Work

Existing methods [9]-[14] were proposed for image generation with applications to colorization (converting a a grayscale image to a color image) [9]-[11] including IR colorization [11], artistic style transfer [12][13], and dataset augmentation [14]. These existing methods are mainly concerned with the generation of new data samples that are similar to existing reference data samples. To achieve this goal, they optimize an objective function that maximizes a similarity measure between the generated data and the reference data. In contrast, our proposed *generative sensing* is based on discriminative models and optimizes a target-oriented objective function (e.g., a regularized categorical cross-entropy loss function) to maximize the classification accuracy rather than a similarity measure.

In [15], we presented a method (*DeepCorrect*) to identify the convolutional filters that are most sensitive to degradations in image quality and to correct the degraded activations of these filters. We showed that correcting only a fraction of the most susceptible filter activations using small low complexity convolutional filter blocks results in a significant performance improvement on popular datasets including ImageNet [16]. We also showed that our *DeepCorrect* model can achieve a classification accuracy higher than fine-tuning (which retrains all network parameters) while only training a fraction of the network parameters, and that it can also train faster than fine-tuning. Our proposed generative sensing framework builds on and extends our *DeepCorrect* work [15] to sensors under varying conditions (e.g., varying illumination, environment, and acquisition characteristics in addition to resolution) and varying modalities (e.g., NIR, IR, Ultrasound, RADAR, in addition to visible wavelength).

## 3. Proposed Generative Sensing

A block diagram of the proposed *generative sensing* framework is shown in Fig. 2. In the following description, the term high-end (low-end) sensor data refers to sensor data that result in a relatively high (low) classification accuracy when using a pre-trained deep neural network (DNN) $\phi$.

During the training phase (Fig. 2a), given high-end sensor data of type X and low-end sensor data of type Y, where types X and Y can be the same (X=Y) or different (X≠Y), feature maps are obtained by applying DNN $\phi$ to each set of data separately. Let $\Phi_{high}$ and $\Phi_{low}$ denote, respectively, the set of feature maps resulting from the high-end sensor data and low-end sensor data. Using a distance measure to quantify differences between co-located features in $\Phi_{high}$ and $\Phi_{low}$, feature difference maps $\Delta\Phi$ are obtained. We are mainly interested in locating feature differences that result in a significant drop in classification accuracy (e.g., a drop greater than a certain user-defined value) by finding features in $\Phi_{low}$ that are *significantly* different from co-located features in $\Phi_{high}$ based on the change in classification accuracy. One approach for locating such features is by measuring the classification accuracy drop when a feature in $\Phi_{high}$ is replaced by its co-located $\Phi_{low}$ feature while keeping all other features in $\Phi_{high}$ unchanged [15]. Another approach for determining significant feature differences can be based on maximizing the drop in classification accuracy by replacing clusters of features in $\Phi_{high}$ with clusters of features in $\Phi_{low}$. Classification-based feature difference maps $\Delta\Phi$ can thus be obtained with values corresponding to the classification accuracy drop. $\Delta\Phi$ can also be thresholded to produce a binary significance difference tensor (significance mask) with a value of 1 denoting a significant feature difference at a location and a value of 0 denoting an insignificant difference. Selective feature regeneration is performed by learning transformations to be applied only to features in $\Phi_{low}$ corresponding to significant $\Delta\Phi$ values, while leaving all other features in $\Phi_{low}$ unchanged. Such transformations can be learned by using relatively small residual learning units as in [15] or by adopting other learning models. We refer to such transformations as *generative transformation units* or simply as *generative units* for short. The resulting *generative sensing* network $\phi_{gen}$ consists of the original pre-trained DNN $\phi$ augmented with generative units that are applied to a select limited number of significant features. During the testing phase (Fig. 2b), low-end type Y sensor data is classified using the generative sensing DNN $\phi_{gen}$ to obtain an improved classification accuracy. In this way, our proposed *generative sensing* aims to convert the low-end sensor into a high-end sensor in terms of matched classification accuracy.

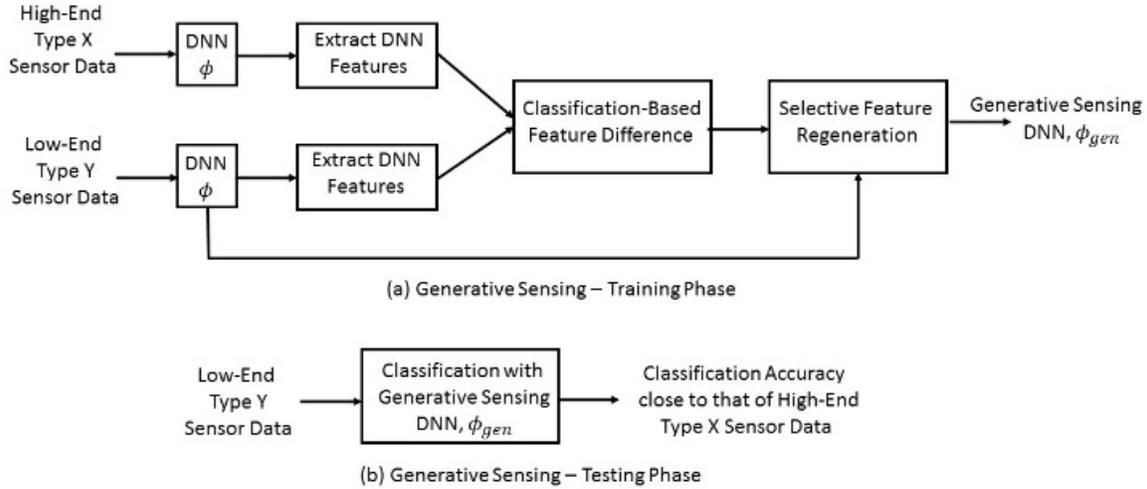

**Figure 2. Proposed Generative Sensing Framework.**

When the generative unit takes the form of a multi-layer network with trainable parameters, the generative transform can be estimated by determining the trainable parameters $\mathbf{W}_{gen}$ of each generative unit so as to minimize a target-oriented loss function such as [15]:

$$E(\mathbf{W}_{gen}) = \lambda\, \rho(\mathbf{W}_{gen}) + \frac{1}{M}\sum_{i=1}^{M} \mathcal{L}(\phi_{gen}(\mathbf{x}_i), y_i)$$

where $\mathcal{L}(.,.)$ is a classification loss function, $y_i$ is a target output label for input $\mathbf{x}_i$, $\phi_{gen}(.)$ is the output of the network with the selectively applied generative units, which we refer to as generative sensing network, $\rho(.)$ Is a regularization term (e.g., $\ell_1$ or $\ell_2$ norm), $\lambda$ is a regularization parameter, and $M$ is the total number of data samples in the training set. The generative units are designed to have a relatively very small number of parameters as compared to DNN $\phi$.

## 4. Performance Results

In order to show the performance of our proposed *generative sensing* framework and its ability to generalize to different tasks (e.g., face recognition, scene recognition), different input modalities (e.g., RGB, NIR, and IR), and different sensor sizes/resolutions, we adopt for the baseline pre-trained DNN $\phi$ of Fig. 2 the baseline AlexNet [4] that has been pre-trained for object recognition on the relatively high-quality visible-wavelength (RGB) ImageNet dataset [16]. For varying tasks and input modalities, we make use of the RGB-IR SCface face recognition dataset [17] and the EPFL RGB-NIR Scene recognition dataset [18]. Unlike the task of face recognition, where the aim is to assign the face in a test image to one of the known subjects in a database, the goal of scene recognition is to classify the entire scene of the image.

The SCface dataset [17] is primarily designed for surveillance-based face recognition. It consists of images acquired in the visible (RGB) as well as infrared spectrum (IR) for 130 subjects, with one frontal mugshot image per subject, for each input modality in addition to other non-frontal mugshots. We make use of the frontal mugshot images. The EPFL RGB-NIR Scene dataset consists of 9 scene categories with at least 50 images per class, for both visible (RGB) and near-infrared spectra (NIR) [18].

We simulate the effect of decreasing the input sensor resolution by blurring the original images by a Gaussian kernel, where the size of the blur kernel is set to 4 times the blur standard deviation $\sigma_b$. Using a blur standard deviation $\sigma_b \in \{0, 1, 2, 3, 4, 5, 6\}$ for Gaussian blur, we simulate 7 successively decreasing levels of sensor resolution, with $\sigma_b = 0$ representing the original high resolution sensor and $\sigma_b = 6$ representing the lowest sensor resolution.

We use the output of the second fully-connected layer in the baseline AlexNet model as a deep feature extractor for the face recognition and scene recognition tasks with images acquired under different sensor resolutions (high to low) and modalities (RGB, NIR, IR). Since the baseline AlexNet model has not been trained on the SCface [17] and EPFL RGB-NIR Scene dataset [18], for each dataset and for each modality, we train a new fully connected layer with a softmax activation on top of deep features that are extracted only from the original dataset images (i.e., $\sigma_b = 0$) and evaluate the performance of such a deep feature extractor for different levels of sensor resolution (i.e., $\sigma_b \in \{0, 1, 2, 3, 4, 5, 6\}$) and for each modality. The new fully connected layer acts like a linear classifier and has a dimensionality equal to the number of classes in the corresponding dataset.

**Table 1:** Top-1 accuracy for images of the SCface dataset [17] acquired under different sensor resolutions and different input modalities, respectively. For each modality (RGB and IR), bold numbers show best accuracy for each resolution level.

| Method-Modality | Sensor Resolution: Blur Level | | | | | | | Avg |
|---|---|---|---|---|---|---|---|---|
| | $\sigma_b = 0$ | $\sigma_b = 1$ | $\sigma_b = 2$ | $\sigma_b = 3$ | $\sigma_b = 4$ | $\sigma_b = 5$ | $\sigma_b = 6$ | |
| Baseline-RGB | **0.9923** | 0.7538 | 0.4384 | 0.3230 | 0.1461 | 0.1000 | 0.0770 | 0.4043 |
| *Generative Sensing*-RGB | 0.9538 | **0.9461** | **0.9000** | **0.8692** | **0.7692** | **0.6846** | **0.6461** | **0.8241** |
| Baseline-IR | **0.9769** | 0.7923 | 0.4769 | 0.1076 | 0.0461 | 0.0076 | 0.0076 | 0.3450 |
| *Generative Sensing*-IR | 0.9000 | **0.8777** | **0.8077** | **0.7538** | **0.6538** | **0.5077** | **0.4692** | **0.7098** |

**Table 2:** Top-1 accuracy for images of the RGB-NIR Scene dataset [18] acquired under different sensor resolutions and different input modalities, respectively. For each modality (RGB and NIR), bold numbers show best accuracy for each resolution level.

| Method-Modality | Sensor Resolution: Blur Level | | | | | | | Avg |
|---|---|---|---|---|---|---|---|---|
| | $\sigma_b = 0$ | $\sigma_b = 1$ | $\sigma_b = 2$ | $\sigma_b = 3$ | $\sigma_b = 4$ | $\sigma_b = 5$ | $\sigma_b = 6$ | |
| Baseline-RGB | **0.9444** | 0.8466 | 0.7644 | 0.6177 | 0.4622 | 0.3511 | 0.2911 | 0.6110 |
| *Generative Sensing*-RGB | 0.9333 | **0.8555** | **0.8511** | **0.8555** | **0.8333** | **0.8355** | **0.8200** | **0.8548** |
| Baseline-NIR | **0.7629** | 0.6733 | 0.5911 | 0.4000 | 0.3088 | 0.2488 | 0.2200 | 0.4578 |
| *Generative Sensing*-NIR | 0.7518 | **0.7200** | **0.7177** | **0.6977** | **0.6622** | **0.6377** | **0.6222** | **0.6870** |

Next we replace the baseline AlexNet DNN $\phi$ (layers 1 to 7) with our generative sensing DNN $\phi_{gen}$ (i.e., baseline AlexNet augmented with generative units) while using the same previously learnt final fully connected layer with softmax activation on top of it as a linear classifier. The generative units were only trained using the ImageNet dataset [16] augmented by adding degraded versions of the ImageNet images with varying levels of blur to simulate varying sensor resolutions [15]. From the results shown in Tables 1 and 2 it can be concluded that, although no images from the SCface [17] and RGB-NIR Scene [18] datasets were used for training, the produced generative units generalize well.

Tables 1 and 2 present the performance results on all levels of sensor resolution for the SCface dataset [17] and the RGB-NIR Scene dataset [18], respectively. It should be noted that we do not learn a new classifier to act on top of our *generative sensing* deep feature extractor, but instead just use the one trained for the baseline AlexNet deep feature extractor.

As shown in Tables. 1 and 2, for both the visible spectrum and the near-infrared/infrared spectrum, the sensor resolution significantly affects the accuracy of the baseline feature extractor, with a 59% and 64% drop in respective average accuracies for the SCface dataset [17], a 35% and 40% drop in respective average accuracies for the RGB-NIR Scene dataset [18] relative to the original sensor resolution (i.e., $\sigma_b = 0$).

For the SCface dataset [17], the *generative sensing* feature extractor significantly outperforms the baseline feature extractor with a 103% and 105% relative improvement in mean accuracy for the visible spectrum and infrared spectrum, respectively. Similarly, for the RGB-NIR Scene dataset [18], the *generative sensing* feature extractor significantly outperforms the baseline feature extractor with a 40% and 50% relative improvement in mean accuracy for the visible spectrum and the near-infrared spectrum, respectively. The large performance gap between the *generative sensing* feature extractor and the baseline AlexNet feature extractor highlights the generic nature of our modality-invariant and sensor resolution-invariant features that are learnt by our *generative sensing* models.

## 5. Conclusion

This work presents a deep learning based *generative sensing* framework for attaining increased classification accuracy on par with the classification accuracy of a high-end sensor while only using a low-end sensor. The high- and low-end sensors can be of different types, in which case the proposed framework can be seen as transforming one type of sensor (e.g., NIR or IR sensor) into another type of sensor (e.g., visible spectrum image sensor) in terms of classification performance. This is achieved through learned transformations that perform selective feature regeneration for improved classification. This is important for enabling robust low-power and low-cost sensing platforms that can work under varying conditions without compromising the recognition performance.

## 6. References

[1] S. Dodge and L. Karam, "Understanding how image quality affects deep neural networks," *International Conference on Quality of Multimedia Experience (QoMEX)*, 6 pages, 2016.

[2] S. Dodge and L. Karam, "A study and comparison of human and deep learning recognition performance under visual distortions," *International Conference on Computer Communications and Networks*, 7 pages, Jul.-Aug. 2017.

[3] S. Dodge and L. Karam, "Can the early human visual system compete with deep neural networks?" *ICCV Workshop on Mutual Benefits of Cognitive and Computer Vision*, 7 pages, October 2017.

[4] A. Krizhevsky, I. Sutskever, and G. E. Hinton, "ImageNet classification with deep convolutional neural networks," *Advances in Neural Information Processing Systems*, pp. 1097–1105, 2012.

[5] E. Fossum, "CMOS image sensors: Electronic camera-on-a-chip," *IEEE Transactions on Electron Devices*, vol. 44, no. 10, pp. 1689-1698, 1997.

[6] M. Bigas, E. Cabruja, J. Forest, and J. Salvi., "Review of CMOS image sensors," *Microelectronics Journal,* vol. 37, no. 5, pp. 433-451, 2006.

[7] J. Sobrino, F. Del Frate, and M. Drusch, "Review of thermal infrared applications and requirements for future high-resolution sensors," *IEEE Transactions on Geoscience and Remote Sensing,* vol. 54, no. 5, pp. 2963-2972, 2016.

[8] A. Rogalski, P. Martyniuk, and M. Kopytko, "Challenges of small-pixel infrared detectors: a review," *Reports on Progress in Physics*, vol. 79, no. 4, pp. 046501, 2016.

[9] A. Deshpande, J. Rock, and D. Forsyth, "Learning Large-Scale Automatic Image Colorization," *IEEE International Conference on Computer Vision (ICCV)*, pp. 567-575, 2015.

[10] R. Zhang, P. Isola, and A.A. Efros, "Colorful Image Colorization," *European Conference on Computer Vision (ECCV), In: Leibe B., Matas J., Sebe N., Welling M.(eds) Computer Vision – ECCV 2016. Lecture Notes in Computer Science*, vol 9907, pp. 649-666, Springer, Cham.

[11] M. Limmer and H.P.A. Lensch, "Infrared Colorization Using Deep Convolutional Neural Networks," *IEEE International Conference on Machine Learning and Applications (ICMLA)*, pp. 61-68, 2016.

[12] L. A. Gatys, A.S. Ecker, and M. Bethge, "A Neural Algorithm of Artistic Style," *arXiv*:1508.06576 [cs.CV], 2015.

[13] L. A. Gatys, A.S. Ecker, and M. Bethge, "Image Style Transfer Using Convolutional Neural Networks," *IEEE CVPR*, pp. 2414-2423, 2016.

[14] V. V. Kniaza, V. S. Gorbatsevicha, and V. A. Mizginova, "THERMALNET: A Deep Convolutional Network for Synthetic Thermal Image Generation," *2nd International ISPRS Workshop on PSBB*, pp. 41-45, May 2017.

[15] T. Borkar and L. Karam, " DeepCorrect: Correcting DNN models against Image Distortions," *arXiv:*1705.02406 [cs.CV], 2017.

[16] O. Russakovsky, J. Deng, H. Su, J. Krause, S. Satheesh, S. Ma, Z. Huang, A. Karpathy, A. Khosla, M. Bernstein, A. C. Berg, and L. Fei-Fei, "ImageNet Large Scale Visual Recognition Challenge," *International Journal of Computer Vision (IJCV)*, vol. 115, no. 3, pp. 211–252, 2015.

[17] M. Grgic, K. Delac, S. Grgic, SCface - surveillance cameras face database, *Multimedia Tools and Applications Journal*, vol. 51, No. 3, pp. 863-879, February 2011.

[18] M. Brown and S. Süsstrunk, Multispectral SIFT for Scene Category Recognition, *IEEE International Conference on Computer Vision and Pattern Recognition (CVPR)*, pp. 177-184, 2011.